\documentclass[journal]{IEEEtran}

\usepackage{times,amsmath,epsfig}
\usepackage{amsthm, url}
\usepackage{amssymb}
\usepackage{latexsym}
\usepackage{lipsum}
\usepackage{balance}
\usepackage{subfigure}
\usepackage{nicefrac}
\usepackage{booktabs}
\usepackage{multirow}
\usepackage{color}

\def \D {\mathbb{D}}
\def \F {\mathbb{F}}
\def \S {\mathcal{S}}
\def \M {\mathcal{M}}

\begin{document}
\title{First Steps Toward Camera Model Identification with Convolutional Neural Networks}

\author{Luca~Bondi,~\IEEEmembership{Student~Member,~IEEE,}
		Luca~Baroffio,
        David~G\"{u}era,
        Paolo~Bestagini,~\IEEEmembership{Member,~IEEE,}
        Edward~J.~Delp,~\IEEEmembership{Life Fellow,~IEEE,}
        and~Stefano~Tubaro,~\IEEEmembership{Senior~Member,~IEEE}
\thanks{Copyright (c) 2015 IEEE. Personal use of this material is permitted. However, permission to use this material for any other purposes must be obtained from the IEEE by sending a request to pubs-permissions@ieee.org.}
\thanks{L. Bondi, L. Baroffio, P. Bestagini and S. Tubaro are with the Dipartimento di Elettronica, Informazione e Bioingegneria,
	Politecnico di Milano, Piazza Leonardo da Vinci 32, 20133, Milan, Italy
	(e-mail: luca.bondi / luca.baroffio / paolo.bestagini / stefano.tubaro@polimi.it).}
\thanks{D. G\"{u}era and E. J. Delp are with the Video and Image Processing Laboratory, School of Electrical and Computer Engineering, Purdue University, West Lafayette, IN, 47907 USA (e-mail: dgueraco@purdue.edu, ace@ecn.purdue.edu).}
\thanks{This material is based on research sponsored by DARPA and Air Force Research Laboratory (AFRL) under agreement number FA8750-16-2-0173. The U.S. Government is authorized to reproduce and distribute reprints for Governmental purposes notwithstanding any copyright notation thereon. The views and conclusions contained herein are those of the authors and should not be interpreted as necessarily representing the official policies or endorsements, either expressed or implied, of DARPA and Air Force Research Laboratory (AFRL) or the U.S. Government.}}

\markboth{Journal of \LaTeX\ Class Files,~Vol.~14, No.~8, August~2015}%
{Bondi \MakeLowercase{\textit{et al.}}: First Steps Towards Camera Model Identification with Convolutional Neural Networks}

\maketitle

\begin{abstract}
Detecting the camera model used to shoot a picture enables to solve a wide series of forensic problems, from copyright infringement to ownership attribution. For this reason, the forensic community has developed a set of camera model identification algorithms that exploit characteristic traces left on acquired images by the processing pipelines specific of each camera model. In this paper, we investigate a novel approach to solve camera model identification problem. Specifically, we propose a data-driven algorithm based on convolutional neural networks, which learns features characterizing each camera model directly from the acquired pictures. Results on a well-known dataset of 18 camera models show that: (i) the proposed method outperforms up-to-date state-of-the-art algorithms on classification of 64x64 color image patches; (ii) features learned by the proposed network generalize to camera models never used for training.
\end{abstract}


\IEEEpeerreviewmaketitle

\section{Introduction}\label{sec:intro}

Due to the increasing availability of image acquisition devices and multimedia sharing platforms, photos are becoming a pervasive part of our daily life. As downloading, copying, forging and redistributing digital material is becoming easier over the years, some forms of regulation and authenticity verification have become of urgent necessity. To this purpose, the forensic community has developed a wide series of tools to reverse-engineer the history of multimedia objects \cite{Stamm2013, Piva2013}.

Among the problems tackled by the image forensics community, one that is still under the spotlight due to its implications in several applications is camera model identification \cite{Rocha2011,Kirchner2015}. This is, given an image, detect which camera model and brand has been used to shoot it. Solving this problem can help an analyst in pointing out the owner of illicit and controversial material (e.g., pedo-pornographic shots, terroristic act scenes, etc.). Moreover, the analysis carried out on image patches can be used to expose splicing forgeries operated between images coming from different cameras \cite{Cozzolino2014, Gaborini2014}.

Attribution of a picture to a specific camera model in a blind fashion (i.e., not leveraging watermarks introduced at photo inception) is possible exploiting intrinsic artifacts left at shooting time by the acquisition process. The idea is that each camera model implements a series of characteristic complex operations at acquisition time, from focusing light rays through lenses, to interpolation of color channels by proprietary demosaicing filters, to brightness adjustment, and also others. As these operations are non-invertible, they introduce some unique artifacts on the final image. These artifacts are footprints that act as an asset to detect the used camera model.

To this purpose, several camera model identification algorithms have been proposed in the literature \cite{Kirchner2015}. Some of them work by searching for specific traces on images under analysis according to a hypothesized analytical model known a priori (e.g., imaging model \cite{Swaminathan2008}, noise characteristics \cite{Filler2008}, demosaicing strategies \cite{Bayram2005, Milani2014b,  Cao2009}, lens distortion  \cite{Choi2006}, gain histograms \cite{Chen2007a}, dust traces \cite{Dirik2008}, natural image modeling \cite{Thai2014}, etc.). Other methods exploit features mainly capturing statistical image properties paired with machine learning classifiers (e.g., local binary patterns \cite{Xu2012}, co-occurrences \cite{Chen2015a, Marra2015, Tuama2016}, etc.). However, all the aforementioned methods rely on manually defined procedures to expose traces characterizing different camera models.

In this paper we reverse the used paradigm by investigating the possibility of solving camera attribution problem using a data-driven methodology. This means that we aim at learning features that characterize pictures shot with different cameras directly from images, rather than imposing any model or hand-crafted feature recipes. Our approach is fully motivated by other successful deep learning-based algorithms developed for forensics tasks \cite{Buccoli2014, Luo2014, Chen2015, Bayar2016, Xu2016}. In particular, we make use of a Convolutional Neural Network (CNN) \cite{LeCun1998, Bengio2009} to capture camera-specific artifacts in an automated fashion, and Support Vector Machines (SVMs) for classification.

The main advantages of using the proposed approach with respect to other state-of-the-art solutions are highlighted by our experimental campaign carried out on more than 13,000 images belonging to 18 camera models from a well-known dataset \cite{gloe2010dresden}. More specifically: (i) as our algorithm does not rely on any analytical modeling, it is less prone to errors due to simplistic assumptions or model simplifications (e.g., linearizations, etc.); (ii) the proposed method is able to work on small image patches (i.e., $64 \times 64$ pixels) with 93\% of accuracy, thus paving the way to applications such as tampering and splicing localization; (iii) the proposed CNN, trained only once on the Dresden dataset, learns a feature extraction methodology that generalizes well on a set of unknown camera models; (iv) the reduced dimensionality of the extracted feature vectors (i.e., 128 elements) enables the use of not sophisticated classification tools (i.e., linear SVMs).

\section{CNN for Camera Model Identification}\label{sec:algorithm}

\begin{figure}[t]
	\centering
	\includegraphics[width=.95\linewidth]{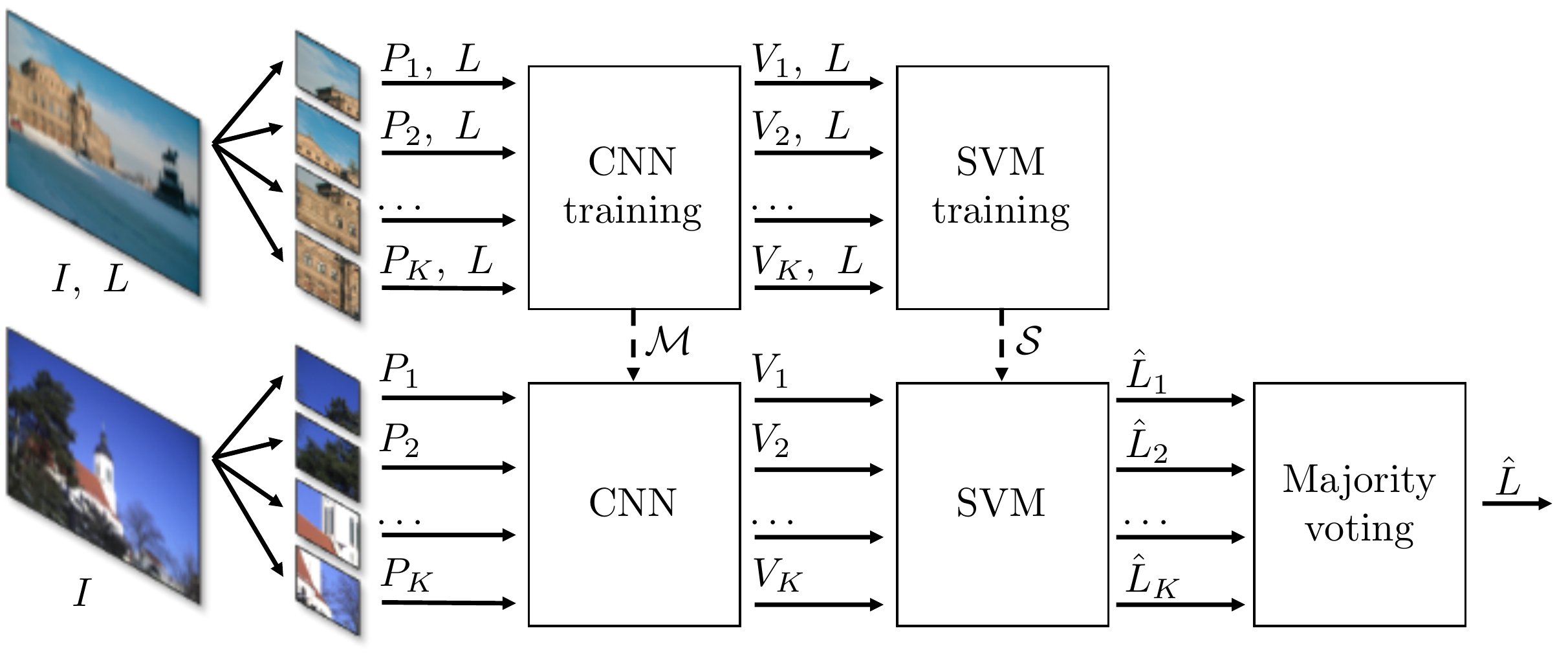}
	\caption{Proposed pipeline for camera model attribution. (Top) CNN and SVM training process: patches extracted from each training image $I$ inherit the same label $L$ of the image. (Bottom) Evaluation pipeline: for each patch $P_k$ from the image $I$ under analysis, a feature vector $V_k$ is extracted through the CNN. Feature vectors are fed to a battery of linear SVM classifiers in order to associate a candidate label $ \hat{L}_k $ to each vector. The predicted label $ \hat{L} $ for image $I$ is obtained by majority voting.}
	\label{fig:pipeline}
\end{figure}

The problem of camera model identification consists in detecting the model $L$ (within a set of known camera models) used to shoot the image $I$. In the following we describe the proposed algorithm to solve this problem\footnote{Code available at \url{https://bitbucket.org/polimi-ispl/}}. We first explain how to perform the training step needed to learn the CNN and SVMs parameters. Then, we report how to use the trained algorithm for classifying new images under analysis. Fig.~\ref{fig:pipeline} shows both training and evaluation pipelines.

\textbf{CNN Training.}
Given a set of training and validation labeled images coming from $N$ known camera models, we train the proposed architecture as follows. For each color image $I$, associated to a specific camera model $L$, we extract $K$ non overlapping patches $P_k, \, k \in [1,K],$ of size $64 \times 64$ pixels. In order to avoid selecting overly dark or saturated regions, we exclude all patches with saturated pixels and prioritize those whose average value is close to half the image dynamic. Each patch $P_k$ inherits the same label $L$ of the source image.
The proposed CNN accepts as input patches of size $64 \times 64 \times 3$, with pixel values between $0$ and $255$. The pixel-wise average over the training set is first subtracted to each input patch. The result is then scaled pixelwise in amplitude by a factor of $0.0125$ to reduce its dynamic. Details about network layers are reported in the following:
\begin{itemize}
	\item A first convolutional layer (\textit{conv1}) with $32$ filters of size $4 \times 4 \times 3$ and stride $1$  is followed by a max-pooling layer (\textit{pool1}) with kernel size $2$ and stride $2$. 
	\item The second convolutional layer (\textit{conv2}) with $48$ filters of size $5 \times 5 \times 32$ and stride $1$  is followed by a max-pooling layer (\textit{pool2}) with kernel size $2$ and stride $2$. 
	\item The third convolutional layer (\textit{conv3}) with $64$ filters of size $5 \times 5 \times 48$ and stride $1$  is followed by a max-pooling layer (\textit{pool3}) with kernel size $2$ and stride $2$.
	\item The fourth convolutional layer (\textit{conv4}) with $128$ filters of size $5 \times 5 \times 64$ and stride $1$  gives as output a vector with $128$ elements.
	\item An inner product layer (\textit{ip1}) with $128$ output neurons is followed by a ReLU layer (\textit{relu1}) to produce a $128$ dimensional feature vector. 
	\item The final $128 \times N$ inner product layer (\textit{ip2}), where $N$ is the number of training classes, is followed by a soft-max layer (\textit{softmax}) for loss computation.
\end{itemize}
The overall architecture is characterized by $340,462$ parameters, learned through Stochastic Gradient Descent on batches of $128$ patches. Momentum is fixed to $0.9$, weights decay is set to $7.5\cdot10^{-3}$ while the learning rate is initialized to $0.015$ and halves every $10$ epochs. As trained CNN model $\M$, we select the one that provides the smallest loss on validation patches within the first $50$ training epochs.

\textbf{SVM Training.}
For each patch, the selected CNN model $\M$ is used to extract a feature vector of $128$ elements, stopping the forward propagation at the \textit{relu1} layer. Feature vectors associated to training patches are used to train a battery of $N\cdot (N-1)/2$ linear binary SVM classifiers $\S$ in a One-versus-One fashion. The regularization constraint $C$ is selected to maximize classification accuracy on validation patches.

\textbf{Deploying the system.}
When a new image $I$ is under analysis, the camera model used to shoot it is estimated as it follows. A set of $K$ patches is obtained from image $I$ as described above. Each patch $P_k$ is processed by CNN model $\M$ in order to extract a feature vector $V_k$. The battery $\S$ of linear SVMs assigns a label $\hat{L}_k$ to each patch. The predicted model $\hat{L}$ for image $I$ is obtained through majority voting on $\hat{L}_k, k\in[1,K]$.

\section{Experimental Results}\label{sec:results}

In order to validate the proposed algorithm, we performed a set of experiments in different operative scenarios: (i) we compared our algorithm against recent state-of-the-art methods; (ii) we evaluated the generalization capability of the proposed CNN. From the implementation side, the CNN training process was managed through the NVIDIA Digits 4.0 Framework, with its built-in Deep Learning Caffe 0.15.9 module~\cite{jia2014caffe}. In the following, we report details about each experiment.

\subsection{Comparison with the State-of-the-Art}
The first experiment aims at validating the proposed approach against state-of-the-art methods. In particular, we selected as benchmark two methods proposed by Chen et al. \cite{Chen2015a} and Marra et al. \cite{Marra2015}. Both of them are based on machine learning pipelines. The former is selected as one of the most recent approaches for classification of full size images. The latter is selected as the algorithm with most promising performances on small image patches. Specifically, within the set of strategies tested in \cite{Marra2015}, we selected the one denoted as SPAM \cite{Fridrich2012}, which outperforms \cite{Xu2012,Celiktutan2008} and \cite{Gloe2012a}.

\begin{figure}[t]
	\centering
	\includegraphics[width=.9\linewidth]{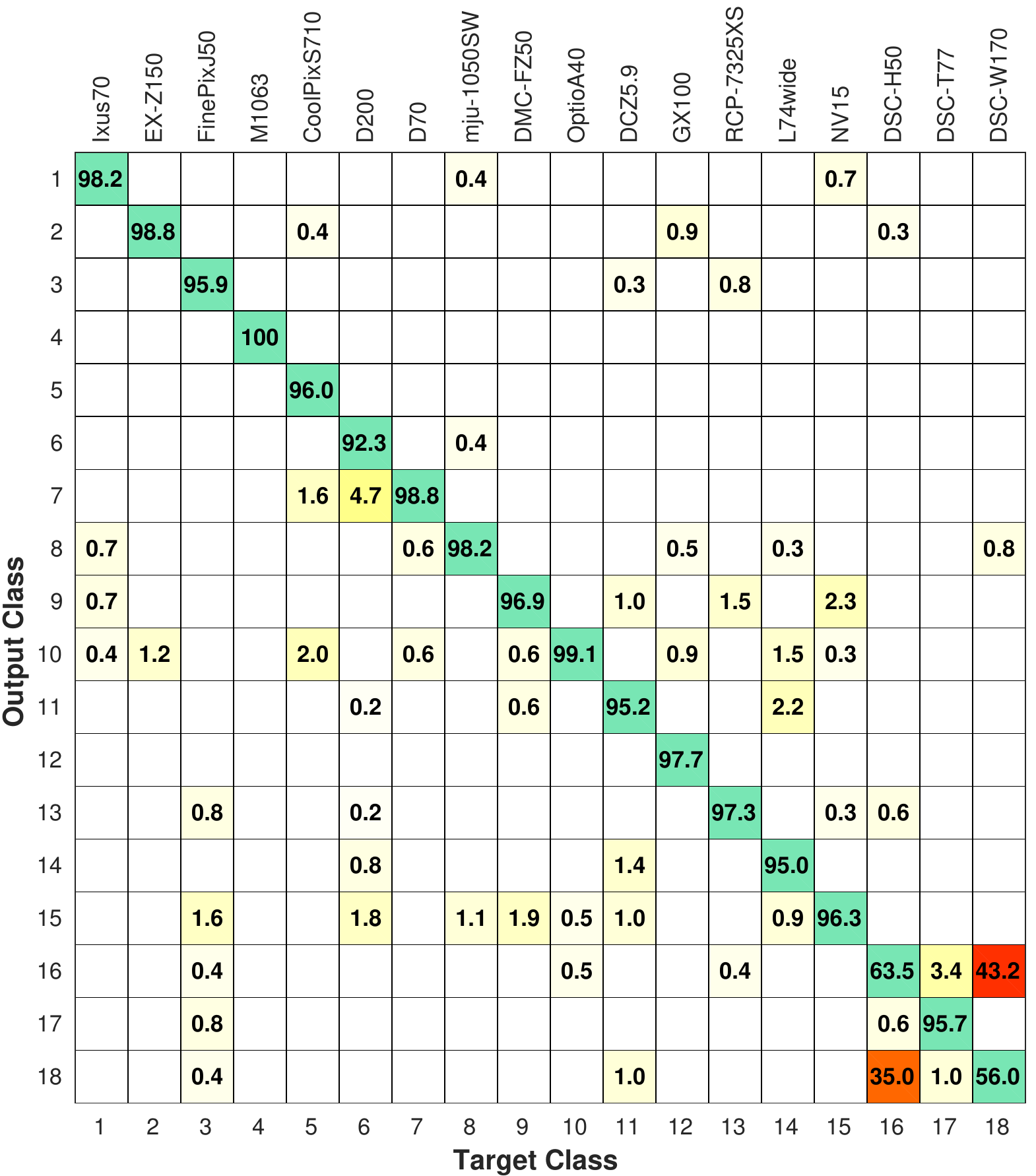}
	\caption{Confusion matrix for the Dresden Image Dataset averaged over 10 splits using the proposed method. Results are obtained with a single patch per image in $\D_E$. Each cell reports the percentage of images from Target Class assigned to Output Class. }
	\label{fig:dresden-cm}
\end{figure}

\textbf{Experimental Setup.}
For this experiment, we considered the Dresden Image Dataset \cite{gloe2010dresden}, which is a publicly available dataset tailored to image source attribution problems. These images are shot with $74$ camera instances of $27$ different models (i.e., several devices of the same model are used).
For each device a variable number of shots has been taken from several locations (e.g., office, public square, etc.). For each location, a set of different pictures was acquired from different viewpoints (e.g., looking on the right, on the let, etc.). Further details about the acquisition process are available at~\cite{gloe2010dresden}. In the following, we will refer to \textit{scene} as the combination of a location and a specific viewpoint.

In this work we selected only natural JPEG photos from camera models having more than one instance. This resulted in $18$ different camera models and $83$ scenes, for a total amount of more than 13,000 images.
Following the idea by Kirchner et al.~\cite{Kirchner2015} we considered the two camera models \textit{Nikon D70} and \textit{Nikon D70s} as a single model.

In order to properly evaluate machine learning based algorithms and ensure a sufficiently large amount of training data, we split the dataset into training, validation and evaluation sets as follows:
\begin{itemize}
	\item we selected evaluation photos ($\D_E$) from $11$ scenes and a single instance per model.
	\item we selected training photos ($\D_T$) from $62$ different scenes and instances.
	\item we selected validation photos ($\D_V$) from the remaining $10$ scenes and instances used for training.
\end{itemize}
As suggested in \cite{Kirchner2015}, this model/scene splitting policy is paramount. Indeed, as scenes and camera instances used during evaluation are never used for training or validation, performance reported on the evaluation set are not biased by scene natural content.

For all tested methods, images from $\D_T$ and $\D_V$ are used to train and tune machine learning tools (i.e., CNN and SVMs for our method, other classifiers for the benchmarks). Evaluation is carried out on images from $\D_E$. Patches are extracted according to what presented in Section~\ref{sec:algorithm} considering $K=32$. The process involves more than $250,000$ training patches from $\D_T$, more than $38,000$ validation patches from $\D_V$, and more than $18,000$ evaluation patches from $\D_E$. All presented results are averaged over 10 cross-validation splits.

\textbf{Results.}
First, we evaluated the accuracy of the proposed method in classifying single image patches (i.e., without voting). To this purpose, Fig.~\ref{fig:dresden-cm} reports the confusion matrix obtained with our method using a single patch per image from $\D_E$. The overall classification accuracy is over $93\%$. As a matter of fact, this is severely affected by the poor performance obtained in separating two \textit{Sony} models, i.e., \textit{DSC-H50} and \textit{DSC-W170},vwhose classification accuracy is respectively of $63.5\%$ and $56.0\%$. It is interesting to notice that these cameras are the ones with the smallest number of shots in the whole dataset and are produced by the same vendor.

\begin{figure}[t]
	\centering
	\includegraphics[width=.85\linewidth]{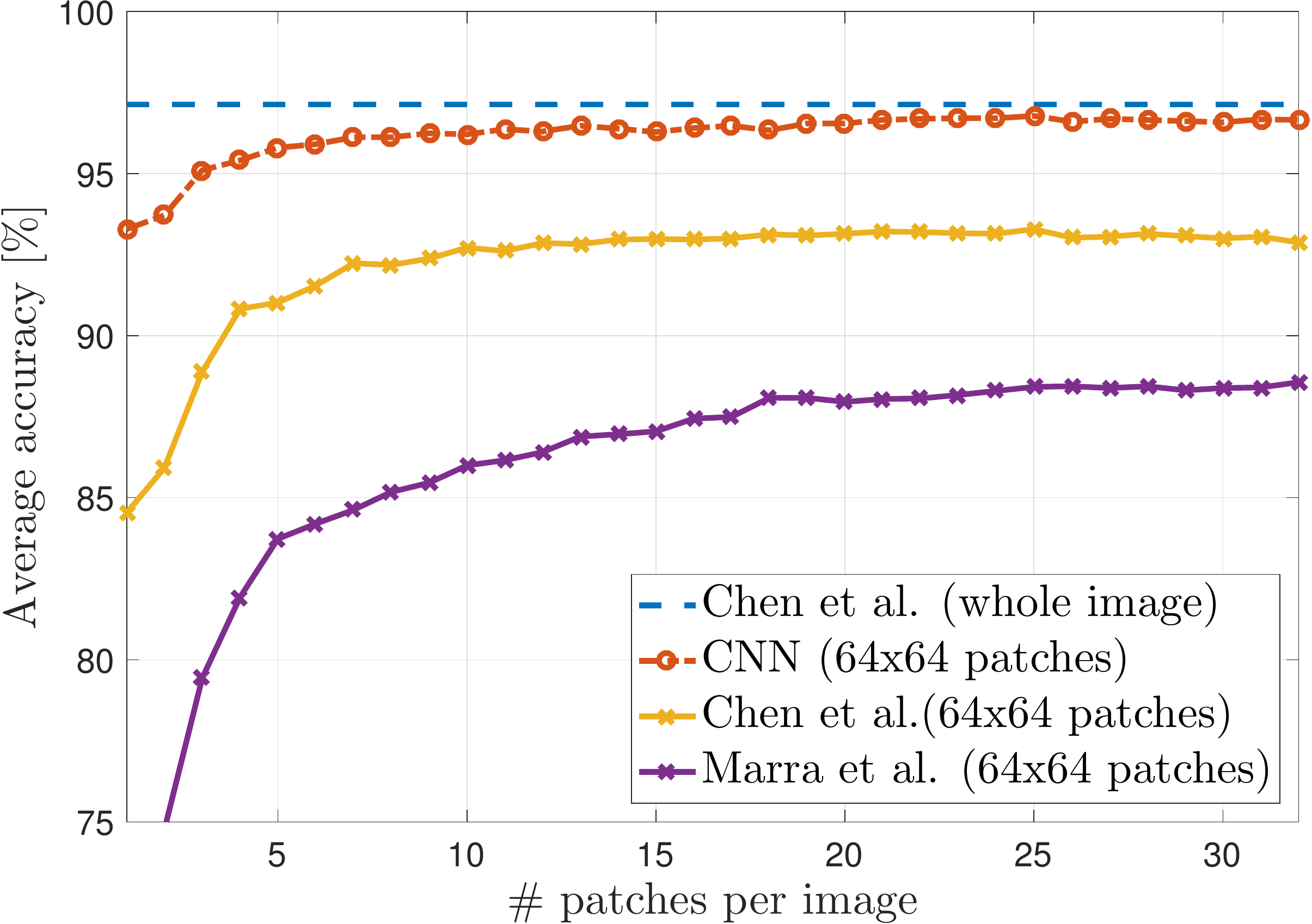}
	\caption{Accuracy for the Dresden Image Dataset while varying the number of patches voting for each image. Accuracy is averaged over 10 splits.}
	\label{fig:ppi_acc}
\end{figure}

After validating the good performance on single patches, we focused on the evaluation of the entire pipeline (i.e., with majority voting) in comparison with state-of-the-art methods. Fig.~\ref{fig:ppi_acc} shows how the average classification accuracy on images from $\D_E$ varies while increasing the number of voting patches for each image. The proposed CNN-based approach is depicted by the orange line. Benchmark results using the approaches proposed by Chen et al.~\cite{Chen2015a} and Marra et al. \cite{Marra2015} applied on $64 \times 64$ color patches followed by majority voting are reported with yellow and purple lines, respectively.

This comparison highlights that our method outperforms all the others on small patches. This is true using either a single patch per image, or majority voting on multiple patches. In particular, the method by Marra et al. suffers from the presence of many camera models, showing worse results than in \cite{Marra2015}.

As the method proposed by Chen et al. is not specifically tailored to small patches, we also tested it on full resolution images without voting procedures (i.e., blue line of Fig.~\ref{fig:ppi_acc}). It is worth noting that, despite the high accuracy obtained by Chen et al., our method approaches within $1\%$ their result by using considerably less input data (i.e., just a few patches and not the full image).

As a final remark, notice that the number of features generated by the CNN for each patch is only $128$, less than one tenth with respect to the $1$,$372$ generated by Chen et al., and less than a half with respect to the $338$ extracted by Marra et al. This means that, even though the CNN as feature extractor is characterized by nearly $340$,$000$ parameters, when it comes to train an external classifier (e.g., SVM) the resulting feature vectors have only $128$ elements. This confirms that we are able to characterize camera models in a space with reduced dimensionality. In principle, this enables the use of simple classifiers, which can be trained more efficiently.

\vspace{-.5em}
\subsection{CNN Generalization Capability.}
One of the seeming negative aspects of the proposed method with respect to the state-of-the-art is that our feature extractor (i.e., the CNN) needs to be trained. Conversely, the other methods rely on manually defined feature extraction techniques. Therefore, in principle, one may think that the proposed method needs to be trained from scratch every time new camera models are considered, thus increasing the computational burden.

Our second experiment aims at disproving this preconception by showing that the feature extraction procedure learned by the CNN well generalizes to novel camera models. As a matter of fact, when new camera models are considered, only SVMs needs to be re-trained (as it happens for classifiers used in \cite{Chen2015a} and \cite{Marra2015}).

\begin{figure}[t]
	\centering
	\includegraphics[width=.5\linewidth]{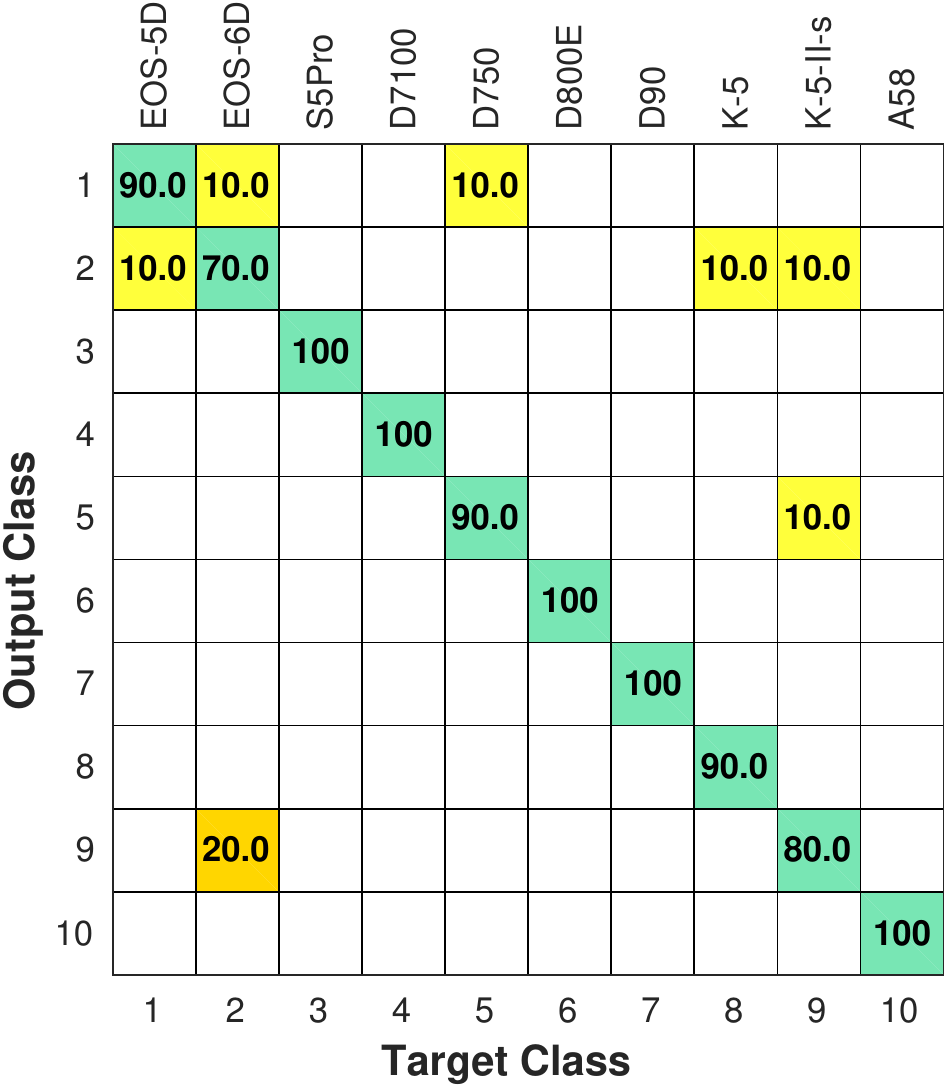}
	\caption{Confusion matrix for the 10-models Flickr Dataset. Results are obtained by voting with 32 patches per image. Each cell reports the percentage of images from Target Class assigned to Output Class.}
	\label{fig:flickr-cm}
\end{figure}

\textbf{Experimental Setup.}
For this experiment, we built a dataset denoted as 10-models Flickr Dataset. This was created by collecting $20$ photos at full resolution from $10$ different camera models (not present in the Dresden Image Dataset) from Flickr website. The dataset is split into a training set $\F_T$ with $10$ shots per model, and an evaluation set $\F_E$ with the remaining $10$ shots per model.

For feature extraction, we selected one CNN model $\M$ trained during the previous experiment on the Dresden Image Dataset (i.e., $\D_T$ and $\D_V$), and we adopted it on patches extracted from images in $\F_T$ and $\F_E$ (i.e., models never seen before by the CNN). A battery of $45$ linear binary SVMs was trained on feature vectors computed on patches extracted from images in $\F_T$. Image from $\F_E$ were used for evaluation.

\textbf{Results.}
Fig.~\ref{fig:flickr-cm} reports the confusion matrix obtained when evaluating the 10-models Flickr Dataset $\F_E$, using majority voting on 32 patches for each image. The overall accuracy reaches $93\%$, showing that the features extracted with the CNN model $\M$, trained on 18 camera models from the Dresden Image Dataset, are capable of generalizing to unknown camera models. As a matter of fact, this result suggests that the CNN learned a feature extraction procedure that is independent from the used camera models. The learned set of operations turned out to be a good procedure to expose traces characterizing different camera models. This confirms that the trained network can be used as other hand-crafted approaches \cite{Chen2015a,Marra2015} not requiring to be trained from scratch every time.

\begin{figure}[t]
	\centering
	\includegraphics[width=.95\linewidth]{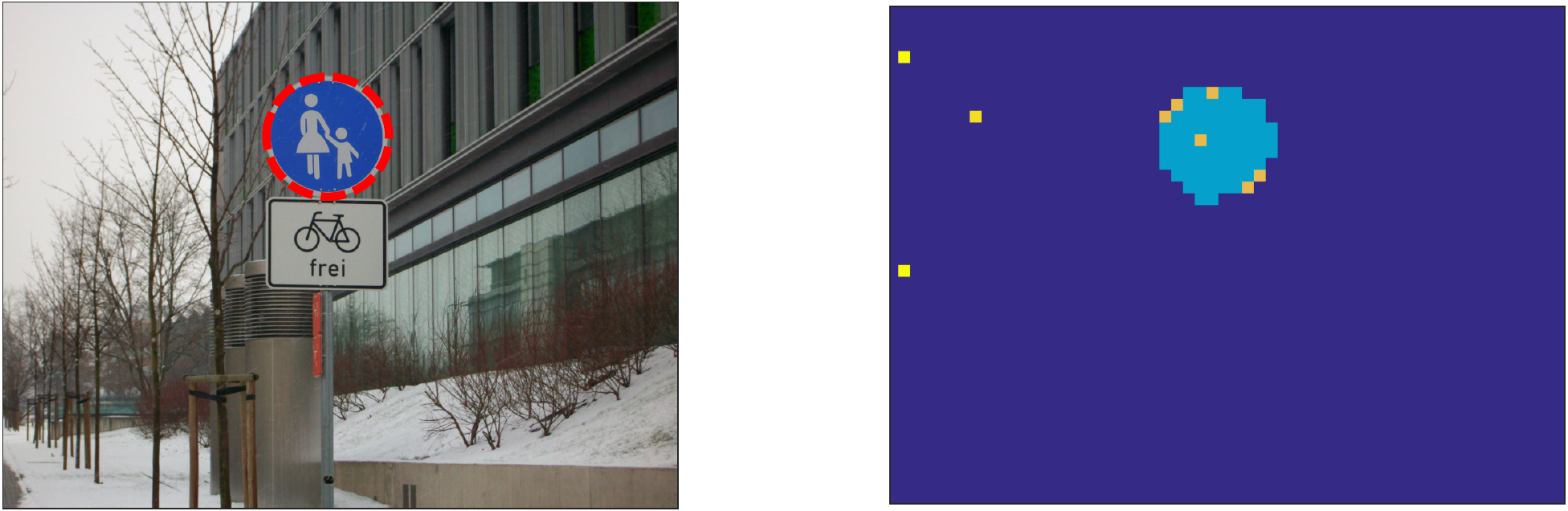}
	\caption{Tampering localization example. We spliced two images from different camera models (inside and outside the red circle). By classifying each block separately and attributing a color to each detected camera model, it is possible to expose the splicing despite a few misclassified blocks (in yellow).}
	\label{fig:tampering}
\end{figure}

\section{Conclusions}\label{sec:conclusions}

In this paper we presented the possibility of using CNNs to solve camera model identification problems. The proposed algorithm has been validated paying particular attention to the used evaluation protocol in order to avoid overfitted or biased results. Comparison against up-to-date state-of-the-art methods show promising results when small image patches are considered. Accuracy achieved in the literature exploiting full size images can be obtained with our method with a considerable reduced amount of information (i.e., a few patches).

Despite being a method based on deep learning, results show that the CNN can be trained only once to learn a feature extraction procedure that can be used afterwards also on images from camera models never seen by the CNN. Moreover, the ability of associating small image patches to camera models, paves the way to the development of image splicing localization methods. To support this intuition, Fig.~\ref{fig:tampering} reports a tampered image in which the background comes from a camera model, and the area marked in red comes from another model. By attributing each image patch to a camera model through the proposed method, it is possible to detect the strong inconsistency between the cameras used to shot the background and the spliced region, respectively.

Additional future work will be devoted to the use of CNNs for camera instance attribution, as a preliminary analysis showed that the proposed technique cannot directly be used to this purpose.

\balance
\bibliographystyle{IEEEtran}
\bibliography{biblio}

\begin{thebibliography}{10}
\providecommand{\url}[1]{#1}
\csname url@samestyle\endcsname
\providecommand{\newblock}{\relax}
\providecommand{\bibinfo}[2]{#2}
\providecommand{\BIBentrySTDinterwordspacing}{\spaceskip=0pt\relax}
\providecommand{\BIBentryALTinterwordstretchfactor}{4}
\providecommand{\BIBentryALTinterwordspacing}{\spaceskip=\fontdimen2\font plus
\BIBentryALTinterwordstretchfactor\fontdimen3\font minus
  \fontdimen4\font\relax}
\providecommand{\BIBforeignlanguage}[2]{{%
\expandafter\ifx\csname l@#1\endcsname\relax
\typeout{** WARNING: IEEEtran.bst: No hyphenation pattern has been}%
\typeout{** loaded for the language `#1'. Using the pattern for}%
\typeout{** the default language instead.}%
\else
\language=\csname l@#1\endcsname
\fi
#2}}
\providecommand{\BIBdecl}{\relax}
\BIBdecl

\bibitem{Stamm2013}
M.~C. Stamm, {Min Wu}, and K.~J.~R. Liu, ``{Information Forensics: An Overview
  of the First Decade},'' \emph{IEEE Access}, vol.~1, pp. 167--200, 2013.

\bibitem{Piva2013}
A.~Piva, ``An overview on image forensics,'' \emph{ISRN Signal Processing},
  vol. 2013, p.~22, 2013.

\bibitem{Rocha2011}
A.~Rocha, W.~Scheirer, T.~Boult, and S.~Goldenstein, ``Vision of the unseen:
  Current trends and challenges in digital image and video forensics,''
  \emph{ACM Computing Surveys (CSUR)}, vol.~43, pp. 26:1--26:42, 2011.

\bibitem{Kirchner2015}
M.~Kirchner and T.~Gloe, ``{Forensic Camera Model Identification},'' in
  \emph{Handbook of Digital Forensics of Multimedia Data and Devices}.\hskip
  1em plus 0.5em minus 0.4em\relax Chichester, UK: John Wiley \& Sons, Ltd,
  2015, pp. 329--374.

\bibitem{Cozzolino2014}
D.~Cozzolino, D.~Gragnaniello, and L.~Verdoliva, ``Image forgery localization
  through the fusion of camera-based, feature-based and pixel-based
  techniques,'' in \emph{IEEE International Conference on Image Processing
  (ICIP)}, 2014.

\bibitem{Gaborini2014}
L.~Gaborini, P.~Bestagini, S.~Milani, M.~Tagliasacchi, and S.~Tubaro,
  ``Multi-clue image tampering localization,'' in \emph{IEEE International
  Workshop on Information Forensics and Security (WIFS)}, 2014.

\bibitem{Swaminathan2008}
A.~Swaminathan, M.~Wu, and K.~J.~R. Liu, ``Digital image forensics via
  intrinsic fingerprints,'' \emph{IEEE Transactions on Information Forensics
  and Security (TIFS)}, vol.~3, pp. 101--117, 2008.

\bibitem{Filler2008}
T.~Filler, J.~Fridrich, and M.~Goljan, ``Using sensor pattern noise for camera
  model identification,'' in \emph{IEEE International Conference on Image
  Processing (ICIP)}, 2008.

\bibitem{Bayram2005}
S.~Bayram, H.~Sencar, N.~Memon, and I.~Avcibas, ``Source camera identification
  based on {CFA} interpolation,'' in \emph{IEEE International Conference on
  Image Processing (ICIP)}, 2005.

\bibitem{Milani2014b}
S.~Milani, P.~Bestagini, M.~Tagliasacchi, and S.~Tubaro, ``Demosaicing strategy
  identification via eigenalgorithms,'' in \emph{IEEE International Conference
  on Acoustics, Speech and Signal Processing (ICASSP)}, 2014.

\bibitem{Cao2009}
H.~Cao and A.~C. Kot, ``Accurate detection of demosaicing regularity for
  digital image forensics,'' \emph{IEEE Transactions on Information Forensics
  and Security (TIFS)}, vol.~4, pp. 899--910, 2009.

\bibitem{Choi2006}
K.~Choi, E.~Lam, and K.~Wong, ``Automatic source camera identification using
  the intrinsic lens radial distortion,'' \emph{Optics Express}, vol.~14, pp.
  11\,551--11\,565, 2006.

\bibitem{Chen2007a}
S.-H. Chen and C.-T. Hsu, ``Source camera identification based on camera gain
  histogram,'' in \emph{IEEE International Conference on Image Processing
  (ICIP)}, 2007.

\bibitem{Dirik2008}
A.~Dirik, H.~Sencar, and N.~Memon, ``Digital single lens reflex camera
  identification from traces of sensor dust,'' \emph{IEEE Transactions on
  Information Forensics and Security (TIFS)}, vol.~3, pp. 539--552, 2008.

\bibitem{Thai2014}
T.~H. Thai, R.~Cogranne, and F.~Retraint, ``Camera model identification based
  on the heteroscedastic noise model,'' \emph{IEEE Transactions on Image
  Processing (TIP)}, vol.~23, pp. 250--263, 2014.

\bibitem{Xu2012}
G.~Xu and Y.~Q. Shi, ``Camera model identification using local binary
  patterns,'' in \emph{IEEE International Conference on Multimedia and Expo
  (ICME)}, 2012.

\bibitem{Chen2015a}
C.~Chen and M.~C. Stamm, ``{Camera model identification framework using an
  ensemble of demosaicing features},'' in \emph{IEEE International Workshop on
  Information Forensics and Security (WIFS)}, 2015.

\bibitem{Marra2015}
F.~Marra, G.~Poggi, C.~Sansone, and L.~Verdoliva, \emph{Evaluation of
  Residual-Based Local Features for Camera Model Identification}.\hskip 1em
  plus 0.5em minus 0.4em\relax Springer International Publishing, 2015, pp.
  11--18.

\bibitem{Tuama2016}
A.~Tuama, F.~Comby, and M.~Chaumont, ``Camera model identification based on
  machine learning approach with high order statistics features,'' in
  \emph{European Signal Processing Conference (EUSIPCO)}, 2016.

\bibitem{Buccoli2014}
M.~Buccoli, P.~Bestagini, M.~Zanoni, A.~Sarti, and S.~Tubaro, ``Unsupervised
  feature learning for bootleg detection using deep learning architectures,''
  in \emph{IEEE International Workshop on Information Forensics and Security
  (WIFS)}, 2014.

\bibitem{Luo2014}
D.~Luo, R.~Yang, and J.~Huang, ``Detecting double compressed {AMR} audio using
  deep learning,'' in \emph{IEEE International Conference on Acoustics, Speech
  and Signal Processing (ICASSP)}, 2014.

\bibitem{Chen2015}
C.~Jiansheng, K.~Xiangui, L.~Ye, and Z.~J. Wang, ``Median filtering forensics
  based on convolutional neural networks,'' \emph{IEEE Signal Processing
  Letters (SPL)}, vol.~22, pp. 1849--1853, 2015.

\bibitem{Bayar2016}
B.~Bayar and M.~C. Stamm, ``A deep learning approach to universal image
  manipulation detection using a new convolutional layer,'' in \emph{ACM
  Workshop on Information Hiding and Multimedia Security (IH\&MMSec)}, 2016.

\bibitem{Xu2016}
G.~Xu, H.~Z. Wu, and Y.~Q. Shi, ``Structural design of convolutional neural
  networks for steganalysis,'' \emph{IEEE Signal Processing Letters (SPL)},
  vol.~23, pp. 708--712, 2016.

\bibitem{LeCun1998}
Y.~Le~Cun and Y.~Bengio, \emph{The Handbook of Brain Theory and Neural
  Networks}, M.~A. Arbib, Ed.\hskip 1em plus 0.5em minus 0.4em\relax MIT Press,
  1998.

\bibitem{Bengio2009}
Y.~Bengio, ``{Learning Deep Architectures for AI},'' \emph{Foundations and
  Trends in Machine Learning}, vol.~2, pp. 1--127, 2009.

\bibitem{gloe2010dresden}
T.~Gloe and R.~B{\"o}hme, ``The {Dresden} image database for benchmarking
  digital image forensics,'' \emph{Journal of Digital Forensic Practice},
  vol.~3, pp. 150--159, 2010.

\bibitem{jia2014caffe}
Y.~Jia, E.~Shelhamer, J.~Donahue, S.~Karayev, J.~Long, R.~Girshick,
  S.~Guadarrama, and T.~Darrell, ``Caffe: Convolutional architecture for fast
  feature embedding,'' \emph{arXiv preprint arXiv:1408.5093}, 2014.

\bibitem{Fridrich2012}
J.~Fridrich and J.~Kodovsky, ``Rich models for steganalysis of digital
  images,'' \emph{IEEE Transactions on Information Forensics and Security
  (TIFS)}, vol.~7, pp. 868--882, 2012.

\bibitem{Celiktutan2008}
O.~Celiktutan, B.~Sankur, and I.~Avcibas, ``Blind identification of source
  cell-phone model,'' \emph{IEEE Transactions on Information Forensics and
  Security (TIFS)}, vol.~3, pp. 553--566, 2008.

\bibitem{Gloe2012a}
T.~Gloe, ``Feature-based forensic camera model identification,'' in \emph{LNCS
  Transactions on Data Hiding and Multimedia Security VIII}.\hskip 1em plus
  0.5em minus 0.4em\relax Springer Berlin Heidelberg, 2012, vol. 7228, pp.
  42--62.

\end{thebibliography}

\end{document}